# An expert system for recommending suitable ornamental fish addition to an aquarium based on aquarium condition

Mohammad Mohammadi
Department of computer science and engineering
Shiraz University
Shiraz, Iran

Shahram Jafari
Department of computer science and engineering
Shiraz University
Shiraz, Iran

**Abstract—** Expert systems prove to be suitable replacement for human experts when human experts are unavailable for different reasons. Various expert system has been developed for wide range of application. Although some expert systems in the field of fishery and aquaculture has been developed but a system that aids user in process of selecting a new addition to their aquarium tank never been designed. This paper proposed an expert system that suggests new addition to an aquarium tank based on current environmental condition of aquarium and currently existing fishes in aquarium. The system suggest the best fit for aquarium condition and most compatible to other fishes in aquarium.

## I. Introduction

### A. Expert systems

#### 1) Expert system
Expert systems are an intelligent computer program that uses knowledge and inference procedures to solve problems that are difficult enough to require significant expertise (Sasikumar, et al. 2007). Expert systems are an intelligent technique for capturing tacit knowledge in a very specific and limited domain of human expertise (Laudon and Laudon 2012).

#### 2) Expert system structure
Expert systems try to mimic expert's way of addressing problems. Expert systems are consist of:

- **Knowledge base:** Part of expert system that contains domain knowledge (Durnkin 1994). Expert systems model human knowledge as a set of rules that collectively are called the knowledge base (Laudon and Laudon 2012). Discovering, extracting, integrating and collecting expert's knowledge into knowledge base is the most important and undertaking task in building an expert system. Knowledge base represents expertise residing in human expert's long term memory.

- **Working memory:** Part of an expert system that contains problem facts that are discovered during session (Durnkin 1994). Working memory contains the facts that we know about current problem. This part of expert system is similar to short memory of a human expert.

- **Inference engine:** Processor in an expert system that matches the facts contained in the working memory with the domain knowledge contained in the knowledge base, to draw conclusions about the problem (Durnkin 1994). Inference engine can work in either backward chaining or forward chaining approach. This part of expert system tries to work similar to human expert reasoning.

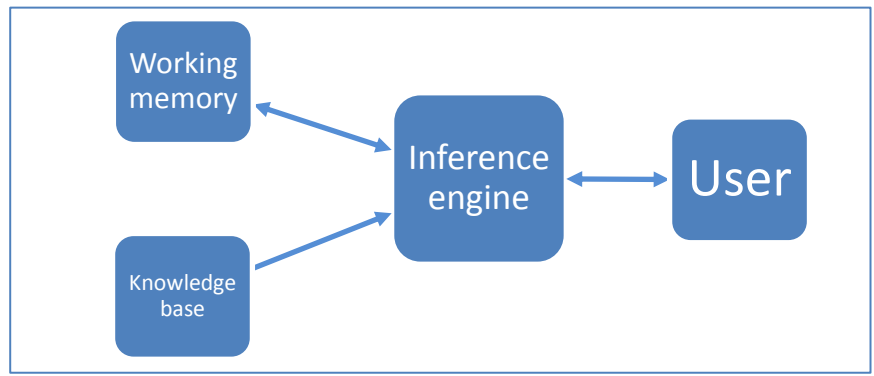

Figure 1: Expert system's structure

- **Explanation facility:** We may want to investigate **how** an expert derived a conclusion or **why** an expert asks a certain question. This can be beneficial to both system developer and system user. The developer can use the explanation facility for debugging. The explanation facility provides transparency to user which leads to user trusting the system or learning from system (Sasikumar, et al. 2007).

*3)* **Expert system's approach**

An expert system in its simplest form works by interacting with user and asking him/her questions about the problem based on its knowledge base. The expert system adds these facts to its working memory. The expert system tries to derive a conclusion by working with its inference engine on the facts from its working memory and knowledge base (Sasikumar, et al. 2007). As explained, expert system's approach of deriving a conclusion is similar to human expert approach.

*4)* **Advantages of expert systems**

Human experts are limited in different aspects. For example a human expert is tied to a certain geographical location and is unavailable in different locations whereas an expert system could be available everywhere without imposing high transportation costs. Table 1 shows advantages of expert systems over human experts.

Table 1: Comparison of human experts and expert systems (Durnkin 1994)

| Factor | Human expert | Expert system |
|---|---|---|
| **Time availability** | Workday | Always |
| **Geographic** | Local | Anywhere |
| **Safety** | Irreplaceable | Replaceable |
| **Perishable** | Yes | No |
| **Performance** | Variable | Consistent |
| **Speed** | Variable | Consistent (Usually faster) |
| **Cost** | High | Affordable |

*5)* **Expert system's applications**

Expert systems are not restricted to certain science area and can be found in almost every field. In (Liao 2005) the author has collected major expert systems from different application and showed expert systems are implementable in vast majority of science areas.

*B. ecessity of proposed system*

*1)* **Importance of aquarium business**

Ornamental fishes are generally aesthetically pleasing fishes that enthusiasts keep in an aquarium tank as a hobby. With about 1.5 and 2 million fans across the world, keeping ornamental fishes is surely a popular hobby. This popularity led to a huge multi-million dollar industry (Wabnitz, et al. 2003). Main market destinations of aquarium keeping industry are United States, the European Union and Japan, respectively (Wabnitz, et al. 2003). Although there is no precise statistics on dimensions of aquarium marketing, there are some estimates emphasizing on significance and profitability of the industry. Volume of traded tropical marine ornamental fishes adds up to 27 million, yearly (Townsend 2011). This huge number of traded fishes makes a 200-330 million dollars' worth industry (Livengood and Chapman 2011).

### 2) Lack of expertise

Aquarium keeping requires extensive amount of experience. Ornamental fishes are generally delicate creatures, very sensitive to their habitat condition. Ornamental fishes must be kept in right condition including water temperature, pH, and hardness. On top of that each fish must be compatible with other tank mates or they may attack and hurt each other. Aquarium enthusiasts, especially amateurs, lack the required expertise for choosing the proper fish for their aquarium tank condition. Poor selection of ornamental fishes would lead to decrease in chance of breeding and early death of them. Death of aquarium fishes not only causes money loss, which is usually considerable, but also instigates frustration feeling in the enthusiast, making him/her disappointed in whole concept of aquarium keeping.

There are three sources that aquarium keepers can use as consultant. But each of these sources bears different problems:

- **Literature:** There are extensive sources of information available in different forms ranging from books, encyclopedias, web pages and forums. Browsing into books and web pages to find information about every fish you may or may not buy and matching its environmental needs against the currently owned fishes as well as their compatibility check can be extremely frustrating. And sometimes impossible due to their unavailability during purchase time. It should be noted that aquarium keeping is a hobby not a chore.
- **Salesperson:** Usually the seller only tries to push buyer into buying more fishes or more expensive fishes, maximizing his or her revenue, no matter if they are the right choice for buyer's aquarium or not. Plus sellers are not necessarily good aquarium keepers. They only have to keep fishes alive for short amount of time usually few days, until they sell the fishes.
- **Experts:** These aquarium keepers possess extensive knowledge about principals of fish keeping. They have learned through experience how each fish must be maintained to maximize its life span and increase the chance of their breeding. The problem with experts is that they may not be available to everyone.

### 3) Suitability of expert system for fish selecting

One of the main applications of expert system is when the expert is not available (Sasikumar, et al. 2007) (Durnkin 1994). As explained in the previous section, in the case of choosing adequate ornamental fish, the expert may not be available. Providing expert knowledge to ornamental fish buyers will help them save money and time as well as keeping them interested in the hobby. Keeping more people interested in the aquarium business would lead to more revenue for fish sellers.

### 4) Objectives

Our system tries suggest best aquarium fish to purchase based on current aquarium condition and fishes. The system firstly takes aquarium condition including water temperature, pH, and hardness and adequate tank size into account. System eliminates inadequate fishes. After determining adequate fishes for aquarium condition, the system gets name of current fishes in the user's tank in order to check their compatibility to the fishes the system want to suggest.

### C. Literature preview

A lot of expert system has been designed in various fields of science. Expert system range from hotel selection (Ngai and Wat 2003) to instruct self-care in case of heart failure (Seto, et al. 2012), predict laser cutting quality (Syn, et al. 2011) and in almost every fields of science. In (Liao 2005) the author summarizes notable expert systems from 1995 until 2004, and mentioned expert systems for various science branches.

Different types of expert system have been designed and implemented in fishery and aquaculture field. In (Deng, et al. 2013) authors suggest an expert system for diagnosing fish diseases. Grant and Berks identified nine categories of knowledge that are important for finding and catching large pelagic fish and then developed an expert system based on this knowledge (Grant and Berkes 2007). Authors in (Cheung, Pitcher and Pauly 2005) focused on a fuzzy expert system that integrates life history and ecological characteristics of marine fishes to estimate their intrinsic vulnerability to fishing.

Although there has been various expert systems and some in fishery and aquaculture field but there has not been a system related to aquarium fishes. The proposed expert system is the first system that helps buyer with selecting adequate ornamental fish.

## II. Designed expert system

Our proposed expert system is a rule based system which has been developed using CLIPS as implementation software. In the following section we explain how we gather and then encode the knowledge needed. We then explain how the system works with this knowledge to suggest addition candidate fishes for an aquarium.

### A. Knowledge acquisition

The first, most undertaking and most important step in developing an expert system is knowledge acquisition (Durnkin 1994) (Sasikumar, et al. 2007). We used different experts toward building our knowledge base. Firstly we used expertise from Encyclopedia of Aquarium & Pond Fish, a well-known encyclopedia (Alderton 2008) and Aquarium Life web site (Pardee n.d.). Secondly we consulted with human experts with both academic and experimental background. These experts work in highly regarded research and education institutions (South of Iran research aquaculture center and fishery department of Islamic Azad University Ahvaz branch) and had themselves experience with aquarium keeping for at least 3 years.

For each fish a profile has been defined to reflect its living condition. We gathered our information from aforementioned sources and experts. Table 2 shows characteristic of a sample fish, namely Molly.

Table 2: Characteristic profile of Molly (Alderton 2008)

| Characteristic | Value |
|---|---|
| **Name** | Molly |
| **Family** | Poeciliidae |
| **Life Span** | 4 years |
| **Minimum Tank Size** | 29 gallons |
| **Temperature Range** | 65°F - 82°F |
| **pH Range** | 7.4 - 8.6 |
| **Hardness Range** | 10° - 30° |

After that we again used aforementioned sources and experts to gather charts showing compatibility of fishes that shows how much a fish is compatible with ther fishes. Figure 2 shows a partial compatibility table with relative compatibility values. As you can see the charts shows how much each fish is compatible with other fishes.

Figure 2: Partial representation of a compatibility chart (Pardee n.d.)

### B. *Knowledge representation*

As we defined the problem, water condition must be checked for every fish. A rule has been defined to implement each constraint on water condition. For example the following rule determines that if water temperature is adequate for fishes in working memory, prompting if the fish is adequate and remove it from working memory. There are similar rules for each affecting water condition factor in knowledge base.

```
(defrule MAIN::check-temp
  (aqua-temp ?temp)
  ?cfish <- (fish (name ?fname) (tempmin ?ftempmin) (tempmax ?ftempmax))
 =>
 (if (> ?ftempmin ?temp)
    then
    (printout t "Your aqua is too cold for " ?fname crlf)
    (retract ?cfish))
 (if (< ?ftempmax ?temp)
    then
    (printout t "Your aqua is too hot for " ?fname crlf)
    (retract ?cfish)))
```

FIGURE 3: SAMPLE RULE FOR ENFORCING TEMPERATURE CONSTRAINT ON EVERY FISH

Since there is not a strict rule for compatibility between two fishes and two fish under certain circumstances may or may not get along, we used certainty factor to undertake this uncertainty. Certainty factor is a number that reflects the net level of belief in a hypothesis given available information (Durnkin 1994). As you can see in Figure 2 some fishes are "sometimes compatible". Different factors such us existence of hiding place for fishes, existence of a specific third fish that gives confidence boost or threatens other fishes, stocking ratio (number of fishes per tank size) and so on can increase or decrease probability of compatibility between two fishes. The affecting factors in fish's compatibility and the way each factor interacts with other factors were determined by experts. In this process new rules has been added to knowledge base to reflect the relation between affecting factors with the regards of certainty factors.

### C. System derivation approach

The expert system's knowledge base contains profile for 100 fishes. Expert starts by placing all these profiles into working memory. Then the system proceeds to asking user about their aquarium condition determining factors affecting the selection. System eliminates inadequate fishes for aquarium condition then presents user with all adequate fishes based on user's tank condition. Then the system asks the user to input all their current fishes. System crossovers these fishes with the fishes in working memory and shows compatible fishes in groups sorted from most compatible to less compatible. Since each candidate fish selected by the system would impose new restriction on future compatible fishes, the results are in form of groups of suggested compatible of fishes not individual fishes.

### III. Results

In order to evaluate the performance of our expert system we compared our expert system with human expert in matter of conclusion derived. We used experience of experts with background of aquarium keeping for at least 3 years from two major education and research institutions, namely, fishery department of Islamic Azad University Ahvaz branch and South of Iran research aquaculture center. We then selected 5 of these experts to compare their advice with our system.

We made test cases of aquarium situation with different aquarium condition and containing different fishes then we asked the expert system and human experts to suggest addition candidate fishes. In average of 80 percent of times, both expert system and expert human derive to same conclusion. We used explanation facility to rationalize the remaining 20 percent of cases. During rationalization process in the average of 15 percent of cases, experts agreed that they forgot one or some affecting factors and the expert system was working correctly.

### IV. Conclusion

We presented an expert system for selecting an adequate candidate fish for adding to an aquarium. There are vast number of different ornamental fishes we focused on the most common aquarium fishes. Different rules has been employed to design the knowledge base of system. We collected information profile for 100 of most common ornamental fishes and added to knowledge base. Certainty factor was used to determine the compatibility of fishes. Results have shown the system's responses are acceptable. This system can help both fish buyers and sellers by helping aquarium enthusiast picking out the right fish for their aquarium.

### V. Suggestion for further research

User of this expert system may want the selection to be based on price of fishes. Prices may vary based on location of user. Even in a certain place, prices may vary during different seasons. A live pricing system could be developed to tackle this issue. Integration of the introduced expert system and a live pricing system could be of great use to aquarium enthusiasts.